\documentclass{article}

     \usepackage[preprint]{neurips_2018}

\usepackage[utf8]{inputenc} 
\usepackage[T1]{fontenc}    
\usepackage{hyperref}       
\usepackage{url}            
\usepackage{booktabs}       
\usepackage{amsfonts}       
\usepackage{nicefrac}       
\usepackage{microtype}      
\usepackage{amsmath,amsthm,amssymb}
\usepackage{graphicx}
\usepackage{wrapfig}
\usepackage{booktabs}

\title{ScieNet: Deep Learning with Spike-assisted Contextual Information Extraction}

\author{
  Xueyuan She \hspace{0.5cm} Yun Long \hspace{0.5cm} Daehyun Kim \hspace{0.5cm} Saibal Mukhopadhyay\\
  Georgia Institute of Technology\\
  Atlanta, Georgia 30332\\
  xshe@gatech.edu \hspace{0.3cm} 
yunlong@gatech.edu \hspace{0.3cm} 
daehyun.kim@gatech.edu \\ saibal.mukhopadhyay@ece.gatech.edu
}

\begin{document}

\maketitle
\vspace{-4mm}
\begin{abstract}
\vspace{-3mm}
 Deep neural networks (DNNs) provide high image classification accuracy, but experience significant performance degradation when perturbation from various sources are present in the input. The lack of resilience to input perturbations makes DNN less reliable for systems interacting with physical world such as autonomous vehicles, robotics, to name a few, where imperfect input is the normal condition. We present a hybrid deep \underline{net}work architecture with \underline{s}pike-assisted \underline{c}ontextual \underline{i}nformation \underline{e}xtraction (\textbf{ScieNet}). ScieNet integrates unsupervised learning using spiking neural network (SNN) for unsupervised contextual information extraction with a back-end DNN trained for classification. The integrated network demonstrates high resilience to input perturbations \textit{without relying on prior training on perturbed inputs}. We demonstrate ScieNet with different back-end DNNs for image classification using CIFAR dataset considering stochastic (noise) and structured (rain) input perturbations. Experimental results demonstrate significant improvement in accuracy on noisy and rainy images \textit{without prior training}, while maintaining state-of-the-art accuracy on clean images.
\end{abstract}

\vspace{-4mm}

\section{Introduction}

Statistical machine learning using deep neural network (DNN) has demonstrated high classification accuracy on complex inputs in many application domains. Motivated by the tremendous success of DNNs in computer vision~\cite{Krizhevsky2012,DBLP:journals/corr/HeZRS15, szegedy2015going} there is a growing interest in employing DNNs in autonomous systems interacting with physical worlds, such as unmanned vehicles and robotics. However, these applications require DNNs to operate on ``novel'' conditions with perturbed inputs. For example, an autonomous vehicle needs to make reliable classifications even with noisy sensor data (stochastic perturbations) or under inclement weather conditions (structured perturbations). 

For convolutional neural networks that depend on local feature extraction, perturbation of pixel level information can cause kernels to map features into incorrect vectors. When such error propagates along the depth of network, resulting classification accuracy is affected~\cite{Nazar2017DeepCN}. It is feasible to tolerate a certain type of perturbations by training the networks on equivalent perturbations, but it is practically intractable to anticipate all possible sources of perturbations during training. Moreover, training to improve accuracy under perturbed images reduces accuracy for clean images~\cite{na2019mixture}. 
Therefore, a new class of DNN architecture that are inherently resilient to input perturbations \textit{without prior training} is necessary for autonomous applications. 

Human brain, on the other hand, can provide accurate classification even under input perturbations. The neuro-inspired computing using Spiking neural network (SNN) has been actively researched for its potential to gain better understanding of biological neural systems and to achieve artificial neural network with learning ability similar to human brain~\cite{Javed2010}. The event based operation principle of SNN~\cite{maass1997networks,izhikevich2008large} differentiates itself from conventional deep learning which utilizes gradient descent. Biologically plausible neuron and synapse models used in SNN makes it possible to exploit temporal relationship between spiking events and optimize network parameters based on causality information~\cite{Moreno-Bote2015, lansdell2019spiking}, which can not be achieved with traditional DNNs. SNN with temporally dependent plastic learning rule has been developed to for image classification purposes. Works have shown network designs for classifying images with both simple~\cite{Diehl2015, She2019FastAndLow, Querlioz2013, Srinivasan2016} and complex ~\cite{tavanaei2018deep, lee2018deep,tavanaei2016bio} features. However, the classification accuracy using unsupervised learning with SNN alone is still far from what is achievable with state-of-the-art DNNs.

This paper presents a hybrid network architecture and learning methodology that couples causal learning of SNN with supervised training based classification using DNN. Our hypothesis is that input perturbation degrades feature quality locally, but at the global level, information from the entire input image is better preserved even under local perturbation. This global level information, if correctly extracted, can be used as contextual information to assist the classification task. We present an implementation of SNN for contextual information extraction and integrate that with a conventional DNN backbone creating a hybrid network. The entire architecture, called ScieNet, achieves more robust image classification under stochastic (noisy) and structured (rainy) perturbation of the input. Specifically, this paper makes the following key contributions:
\begin{itemize}
\item We present a hybrid deep learning architecture with SNN based network components that pre-learn input contextual information without supervision and utilize it in DNN image classification.
\item We demonstrate that the proposed architecture is resilient to input perturbation while requiring no prior knowledge of the perturbation during training and inference. Moreover, as the network is never trained with perturbed inputs, the performance of the proposed network for clean images is not affected. We demonstrate resilience to stochastic (noisy) and structured (rain) input perturbations.  
\item We show that ScieNet is a versatile design that can be easily integrated with different back-end deep learning architectures. 

\end{itemize}

We demonstrate ScieNet with three different back-end DNN architectures, namely, MobileNetV2, ResNet101 and DenseNet, for image classification using the CIFAR10 dataset. The experiments are peformed considering Gaussian noise and rain (synthesized using~\cite{liu2018erase,Garg:2006:PRR:1141911.1141985} and practical) added to the images during inference. All versions of ScieNet shows signifcant improvement in accuracy under noise and rain when compared to the respective backbone DNNs in isolation.

\paragraph{Related work}

The impact of noise and rain on image classification has received significant interest in recent years.  Nazar~\cite{Nazar2017DeepCN} and Luo~\cite{luo2014deep} shows that noise in inference images causes degradation of image classification performance of DNN.

But majority of the past approaches focused on using either additional training with noisy/rainy images to improve accuracy. 
Solutions that have been proposed include training with dataset containing noise~\cite{Nazar2017DeepCN} and manually introducing noise to network parameters~\cite{luo2014deep}. Such approaches are able to improve accuracy when noise pattern used in training is similar to that in inference, but negatively affect network performance on clean dataset. 

A parallel approach is to develop specialized pre-processing network, often using another DNN trained using noisy/rainy images, to de-noise or de-rain the input images before classification. For example, Na et. al. have demonstrated pre-processing network to improve classification accuracy under noise~\cite{na2018noise}.  A recent paper by Liu~\cite{liu2018erase} shows an machine learning approach to remove rain drops from images.

\section{Network Configuration}
\label{sec_network}
\subsection{Spiking neural network}
\paragraph{Spiking neuron model}
Different models have been developed to describe the evolution of membrane potential during SNN simulations at varying fidelity~\cite{Long2018, izhikevich2003simple, Hodgkin1990, rezende2011variational, pfister2006optimal, gardner2015learning}. In this work, we use leaky integrate-and-fire (LIF) model, which consists of one differential equation and one reset function as described by the following equations:
\begin{equation}
dv/dt = a+bv+cI \label{eq: lif_1}\textbf{}
\end{equation}
\begin{equation}
v=v_{reset}, \textit{ if } v>v_{threshold} \label{eq: lif_2}
\end{equation}

Three parameters of \ref{eq: lif_1}, $a$, $b$ and $c$, controls spiking frequency of the neuron model under certain input. Current $I$ to one neuron is the sum of current from all synapses connected to the neuron. More specifically, for neuron $m$ with total income connection of $N$, $I$ is determined by:

\begin{equation}
I_{m} = \sum_{n=0}^{N} g_{n,m}v_{n} \label{eq: lif_3}
\end{equation}

Here, $g_{n,m}$ is the synapse conductance for connection between neuron $n$ and neuron $m$, and $v_{n}$ is the voltage signal sent by neuron $n$. When membrane potential of a spiking neuron crosses threshold $v_{th}$, the neuron generates a spike and its membrane potential is reset to $v_{reset}$. 

\begin{figure}[t]
\centerline{\includegraphics[width=0.9\textwidth]{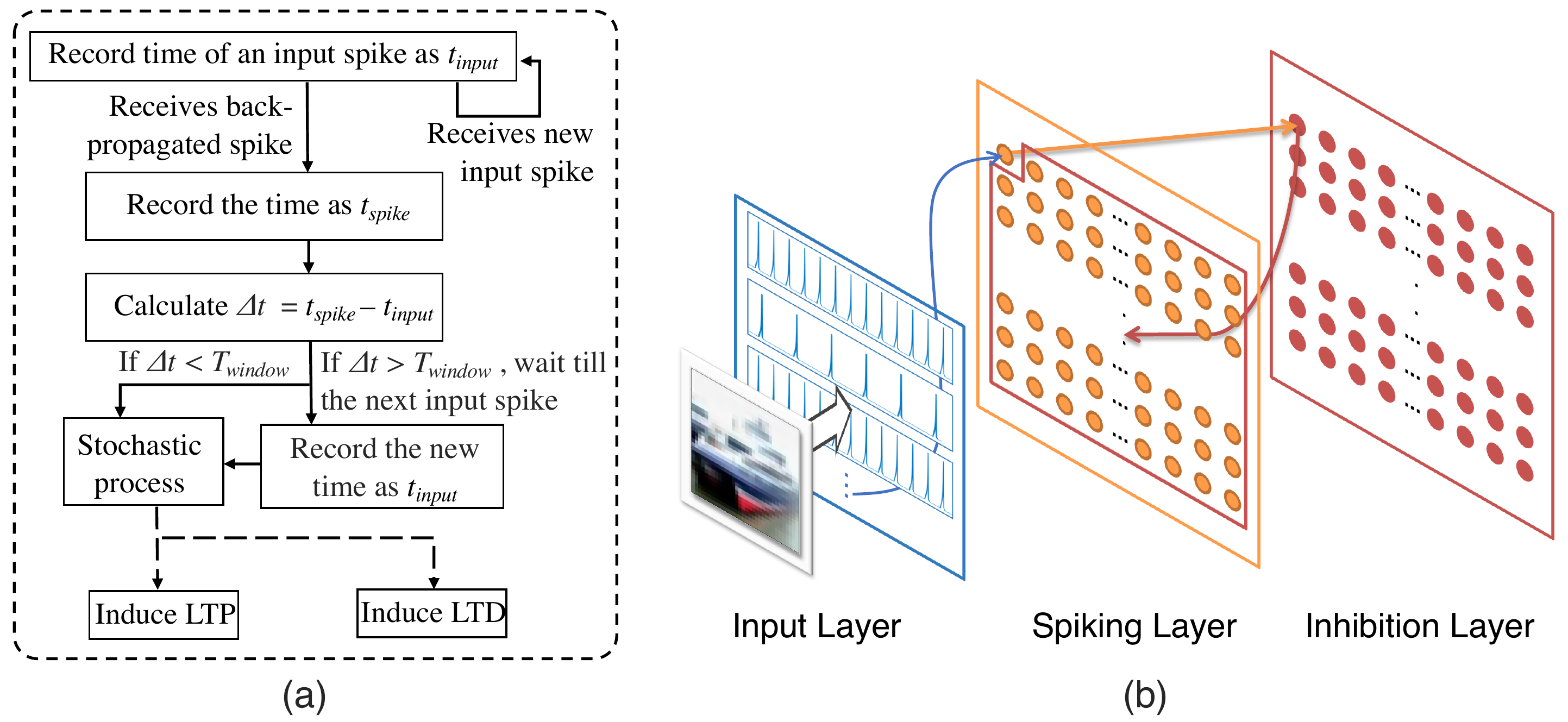}}
\caption{(a) Flowchart of frequency-dependent stochastic STDP learning rule; (b) SNN architecture for contextual inforamtion extraction.}
\label{fig_snn}
\end{figure}

\paragraph{Synapse model}
Spiking-timing-dependent plasticity (STDP)~\cite{fremaux2016neuromodulated, legenstein2008learning} is one type of SNN learning rules that modulate synapse conductance based on specific timing pattern of pre-synaptic and post-synaptic spikes. It governs modulation of synapse conductance with no label required, and is widely adpoted in SNN applications~\cite{Diehl2015, Querlioz2013, Tavanaei2016}. For STDP, long-term potentiation (LTP) and long-term depression (LTD), which increases and decreases synapse conductance respectively, are the two main modulation behaviors. LTP is induced when post-synaptic neuron spikes shortly after receiving an input pre-synaptic spike, indicating a high level of causality between the two events; LTD is induced when pre-synaptic spike is received after post-synaptic neuron spikes. Fig.~\ref{fig_snn}(a) shows the detailed STDP process. When implementing this algorithm, a LTP timing window parameter called $T_{window}$ is used as timing reference. LTP is induced only when post-synaptic neuron spikes during $T_{window}$ following an input spike. Specifically, the modulation of synapse with conductance $G$ is determined by:

\begin{equation}
\Delta G_{p} = \alpha_{p}e^{-\beta_{p}({G-G_{min}})/({G_{max}-G_{min}})} \textbf{}
\label{eq_STDP_1}
\end{equation}
\begin{equation}
\Delta G_{d} = \alpha_{d}e^{-\beta_{d}({G_{max}-G})/({G_{max}-G_{min}})} \textbf{}
\label{eq_STDP_2}
\end{equation}

Here $\Delta G_{p}$ and $\Delta G_{d}$ are the magnitude of modulation; $G_{min}$ and $G_{max}$ are network parameters. The basic STDP algorithm as introduced here captures the general behavior of plastic synapses but does not address the associativity~\cite{Levy1979} issue. The basic STDP algorithm is able to extract features from less complex datasets such as MNIST but falls short for datasets such as CIFAR. In this work we choose to use one type of stochastic STDP algorithm~\cite{Srinivasan2016}, in which conductance modulation is non-deterministic. Specifically, given the spike timing difference $\Delta t$, probability $P$ of conducting the modulation operation is determined by:

\begin{equation}
P_{pot} = \gamma_{pot}e^{(-\Delta t/(\tau_{pot}(1+\Phi_{pot})))}
\label{eq: STOCH_STDP_new_1}\textbf{}
\end{equation}

\begin{equation}
P_{dep} = \gamma_{dep}e^{(\Delta t/(\tau_{dep}(1+\Phi_{dep})))}
\label{eq: STOCH_STDP_new_2}\textbf{}
\end{equation}

\begin{equation}
\Phi_{pot} = \phi_{pot}\frac{f-f_{min}}{f_{max}-f_{min}}
\label{eq: STOCH_STDP_new_5}\textbf{}
\end{equation}

\begin{equation}
\Phi_{dep} = \phi_{dep}\frac{f-f_{min}}{f_{max}-f_{min}}
\label{eq: STOCH_STDP_new_4}\textbf{}
\end{equation}

Here, $f_{min}$ and $f_{max}$ are network parameters that control minimal and maximal frequency of input spikes. When $\Delta t$ is smaller, the probability of a causal relationship existing is higher for potentiation and lower for depression. Correspondingly, value of $P$ for both LTP and LTD is higher. Frequency dependent components $\Phi$ are used in \ref{eq: STOCH_STDP_new_5}, to dynamically adjust the time constant $\tau$ of probability function. Input spike frequency dependence is added to provide account for the associativity issue. The result algorithm allows SNN to work on complex dataset as tested in this work.

\paragraph{SNN architecture}
SNN architecture based on the described neuron model and learning rule can be constructed for the contextual information extraction task. Similar to networks used in~\cite{Srinivasan2016,She2019FastAndLow,Querlioz2013}, a three-level structure as shown in Fig.~\ref{fig_snn}(b) is used. The input layer is an image to spike train converter, as each pixel from each channel in input image is converted to a spike train with specific spiking frequency. The intensity value is scaled to a frequency range from $f_{min}$ to $f_{max}$, with darker pixels having higher spiking frequency. For CIFAR dataset, this creates an array that contains $32\times32\times3=3072$ spike trains.

The second level is an array of LIF neurons each receiving input from all spike trains in the first level, i.e. the first and second level are fully connected. The third level is the inhibition layer which contains inhibitory neurons and the number is equal to that of the spiking neurons in the second level. Each inhibitory neuron has input connection from one corresponding neuron in the spiking neuron layer. When it receives an input spike, inhibitory signal is sent to all other spiking neurons for a period of $t_{inh}$. Membrane potential of spiking neurons that receive inhibitory signal decreases by $\Delta v_{inh}$, and does not accumulate during the inhibition period $t_{inh}$. This creates a lateral inhibition behavior among spiking neurons, which prevents multiple neurons from learning the same input context. In this network, synapses between the first and second layers are plastic, i.e. modulated by STDP, while synapses between the second and third layers are not.

\subsection{ScieNet Architecture}
\begin{figure}[t]
\centerline{\includegraphics[width=1\textwidth]{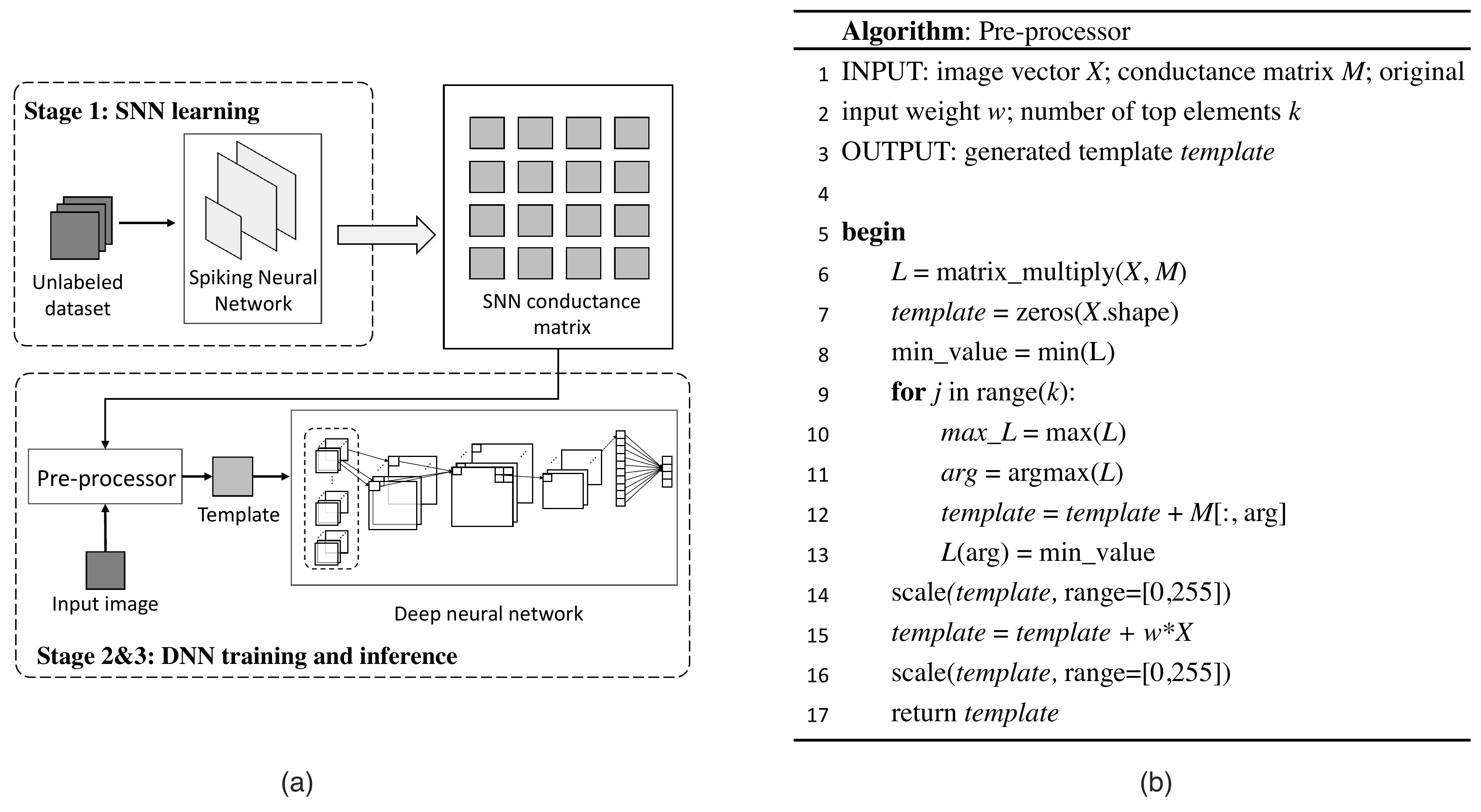}}
\caption{(a) Architecture of ScieNet (b) Algorithm for ScieNet's pre-processor}
\label{fig_SCIE}
\end{figure}

The proposed hybrid architecture of ScieNet is shown in Fig.~\ref{fig_SCIE}(a). Its operation is divided into three stages: unsupervised learning, training and inference. In the first stage, SNN learns the entire training dataset with no labels required. All synapses connected to one neuron form a matrix of conductance that represents the context learned by that neuron. After learning is complete, the matrices of synapse conductance to all spiking neurons are used in the pre-processor for the second stage. The pre-processor as shown in Fig.~\ref{fig_SCIE}(b) utilizes the conductance matrices to generate matching context as detected in the input image and construct synthetic templates based on it. More details of the pre-processor is discussed in the next section. Templates and input labels are then used to train the subsequent deep network. This concludes the training stage of the system. 

During the inference stage test images are sent to the pre-processor for the same template generation procedure and the deep network provides classification results based on the templates. Since the nature of perturbation in images is treated as an unknown property during training and inference, the network is not trained separately for different interference, and parameters of pre-processor and DNN are kept the same across testing in all conditions. All datasets for SNN learning and DNN training consist of clean images only. For the DNN component of the hybrid architecture, we choose to test MobileNetsV2, ResNet and DensNet, to demonstrate that ScieNet can be implemented with different deep network designs. 

\paragraph{Pre-processor details}

Fig. \ref{fig_SCIE}(b) shows the algorithm for the pre-processor. The pre-processor takes as input the conductance matrices $M$, input image vector $X$ and parameters $w$ and $k$. Since the input image could contain a mixture of contextual information learned by each individual spiking neuron. The generated context from pre-processor is taken as a combination of matrices that are selected according to a scoring process. Specifically, for the scoring process, each input image goes through matrix multiplication with the conductance matrix. Here image pixel intensity value is used without conversion to spikes. Therefore, given the $n$ dimensional feature vector $x$ from input image, conductance matrices from $d$ number of spiking neurons acts as an operator $M$ such that:
\begin{equation}
M:\quad {\rm I\!R}^n \rightarrow {\rm I\!R}^d
\label{eq: space_trans}\textbf{}
\end{equation}
For vector in the resulting space $L\in \rm I\!R^d$ and $L=[l(1), ...l(d)]$, top $k$ elements $[l(i_1), ...l(i_k)]$ are selected. Each element in this set has one corresponding conductance matrix from a spiking neuron. The corresponding set of matrices is summed with element-wise addition to generate the context information. The result has a dimension of $n$, the same as input vector $x$. A weighted element-wise addition of context matrix and input vector is calculated, and resulting template is re-scaled then used to train the DNN. Since the dimension of input vector remains the same, the architecture of DNN does not need to be changed to work with ScieNet, making it a versatile and scalable design.

\section{Experimental Results}
\label{sec_result}

\paragraph{Experimental Setup.}
In this work we focus on experimenting with the CIFAR10 dataset. To test the classification accuracy on noisy image, additive white Gaussian noise (AWGN) is applied to the inference images. The noise level is measured in signal-to-noise ratio (SNR). As rainy environment is another frequently encountered condition for real-life applications of deep neural network, we choose to investigate its impact on network performance. To create images affected by rainy condition efficiently, we take into consideration work from Liu~\cite{liu2018erase} and Garg\cite{Garg:2006:PRR:1141911.1141985} when developing a method to add heavy and light rain effect on clean images. In addition to modeled rainy conditions, we also use video footage taken in real-life rainy environment to test classification accuracy of the subject network.

\begin{figure}[]
\centerline{\includegraphics[width=0.95\textwidth]{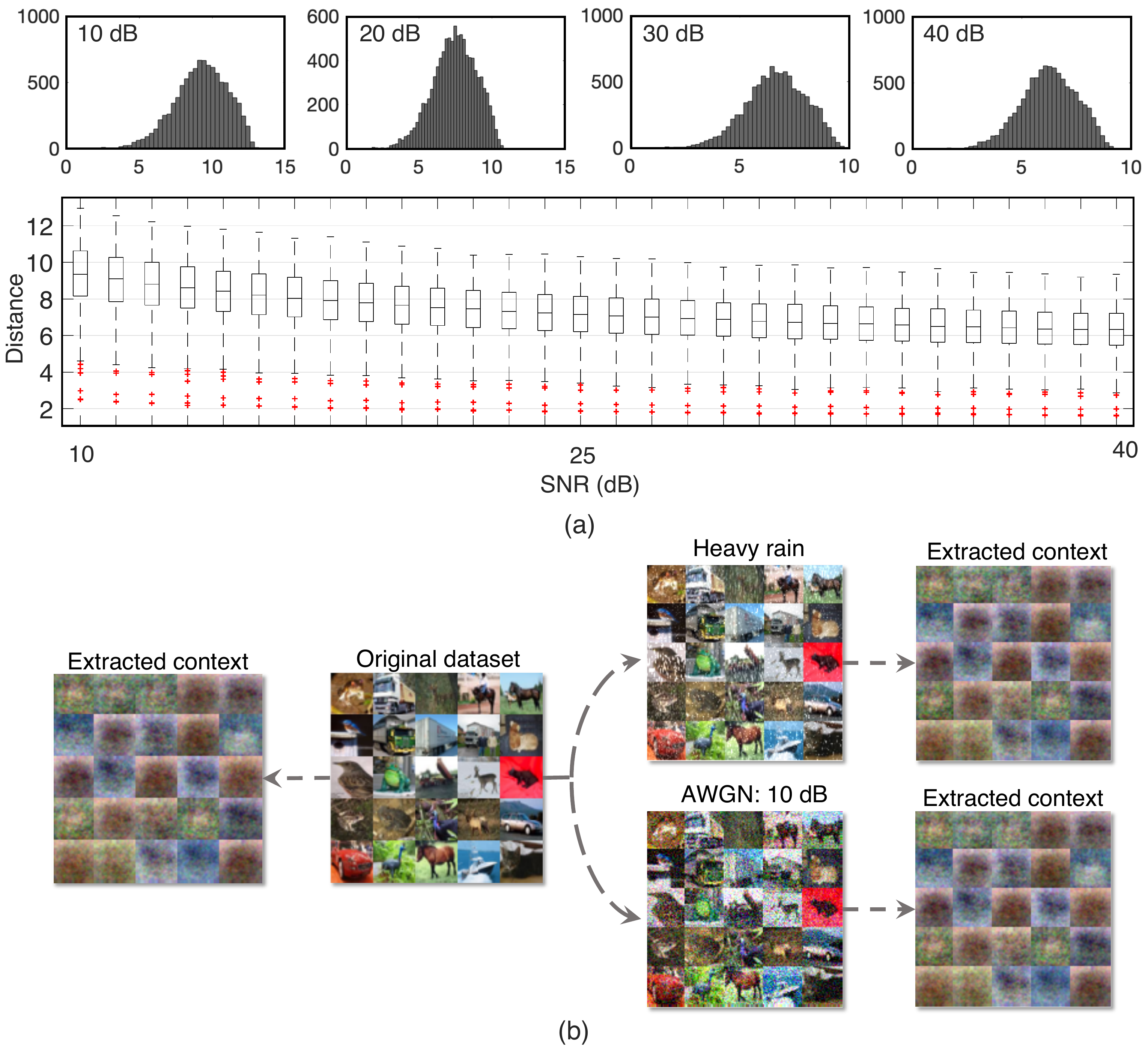}}
\caption{(a) Euclidean distance distribution histogram for context matrix generated for different levels of noise (top) and box plot showing distance variation versus SNR range; (b) example of images with perturbation and extracted context.}
\label{fig_distance}
\end{figure}

\paragraph{Input-Perturbation Resilient Context Extraction using SNN.}
Our key hypothesis is that the contextual information extracted from input image is robust to perturbation. To empirically support this hypothesis, we use the pre-processor for images with different levels of noise and calculate the Euclidean distance between the generated context matrices. The test sample is CIFAR inference set, and results are shown in Fig.\ref{fig_distance} (a). As for the entire test sample space, when SNR decrease from 40 dB to 10 dB the mean of Euclidean distance increase from 6.27 to 9.26. In terms of distribution, the standard deviation increases from 1.28 to 1.80. As can be observed from the example images and extracted context pairs in Fig.\ref{fig_distance} (b), when noise level increases the context matrix stays visually similar. Overall, context matrices remain close in space over the range of noise setting tested. This shows that the context extraction process by pre-processor is not heavily influenced by input perturbations.

\begin{table}
  \caption{Accuracy (\%) results for CIFAR10 with noise}
  \label{table: acc}
  \centering
  \begin{tabular}{llllllll}
    \toprule
    \multicolumn{5}{r}{SNR}                   \\
    \cmidrule(r){3-8}
    Model                   &  Clean & 40 dB & 30 dB & 25 dB & 20 dB &  15 dB & 12 dB\\
    \midrule
    MobileNetV2\cite{DBLP:journals/corr/abs-1801-04381}            & 91.30  & 90.86  & 84.85  & 66.25  & 35.13  & 18.50  & 14.26 \\
    ResNet101\cite{DBLP:journals/corr/HeZRS15}               & 93.57  & 89.74  & 86.39  & 78.32  & 55.47  & 26.33  & 15.53 \\
    DenseNet121\cite{DBLP:journals/corr/HuangLW16a}             & 93.00  & 92.87  & 89.84  & 82.59  & 60.42  & 27.10  & 16.88 \\
    \midrule
    Noise trained MobileNetV2 & 90.54  & 90.64 & 90.16  & 86.36  & 62.22  & 25.37 & 16.51 \\
    Noise trained ResNet101   & 92.41  & 92.51 &  92.26 & 90.92  & 77.81  & 35.97 & 19.85 \\
    Noise trained DenseNet121   & 91.88  & 91.86 & 91.71  & 90.74  & 75.35 & 33.89  & 19.35 \\
    \midrule
    \textbf{ScieNet with MobileNetV2}  & 91.32  & 91.13 & 90.93  & 90.54  & 86.98  & 64.83  & 37.96 \\
    \textbf{ScieNet with ResNet101}    & 93.52  & 93.36 & 93.12  & 92.04  & 89.08  & 68.50  & 41.38 \\
    \textbf{ScieNet with DenseNet121}   & 93.07  & 92.94 & 92.57  & 91.90  & 87.26  & 60.47  & 31.49 \\
    \bottomrule
  \end{tabular}
\end{table}

\paragraph{Accuracy of ScieNet with Noisy images}
Table.\ref{table: acc} compares accuracy results for three test categories: baseline DNN, networks trained with 30 dB noisy images, and ScieNet. For the three baseline networks, noise in images exhibits a significant impact on classification accuracy, as their performance degrades from over 90\% to just above random guessing for 12 dB SNR. Networks trained with noise shows better accuracy under noisy input. The improvement is more prominent when inference noise is at similar level with training noise, but for clean images the performance is negatively affected. 

For ScieNet implemented with all three DNNs as back-end, performance in noisy condition is improved over baseline by a large margin. At 20 dB SNR ScieNet with MobileNetV2 remains at good performance while the baseline drops below 40\% accuracy, making a significant 50\% gain. Compared to networks trained with noise, ScieNet shows similar performance at around 30 dB SNR while its advantage increases at higher noise levels. At the same time, for clean images accuracy of ScieNet is on par with baseline networks.

\begin{figure}[]
\centerline{\includegraphics[width=1\textwidth]{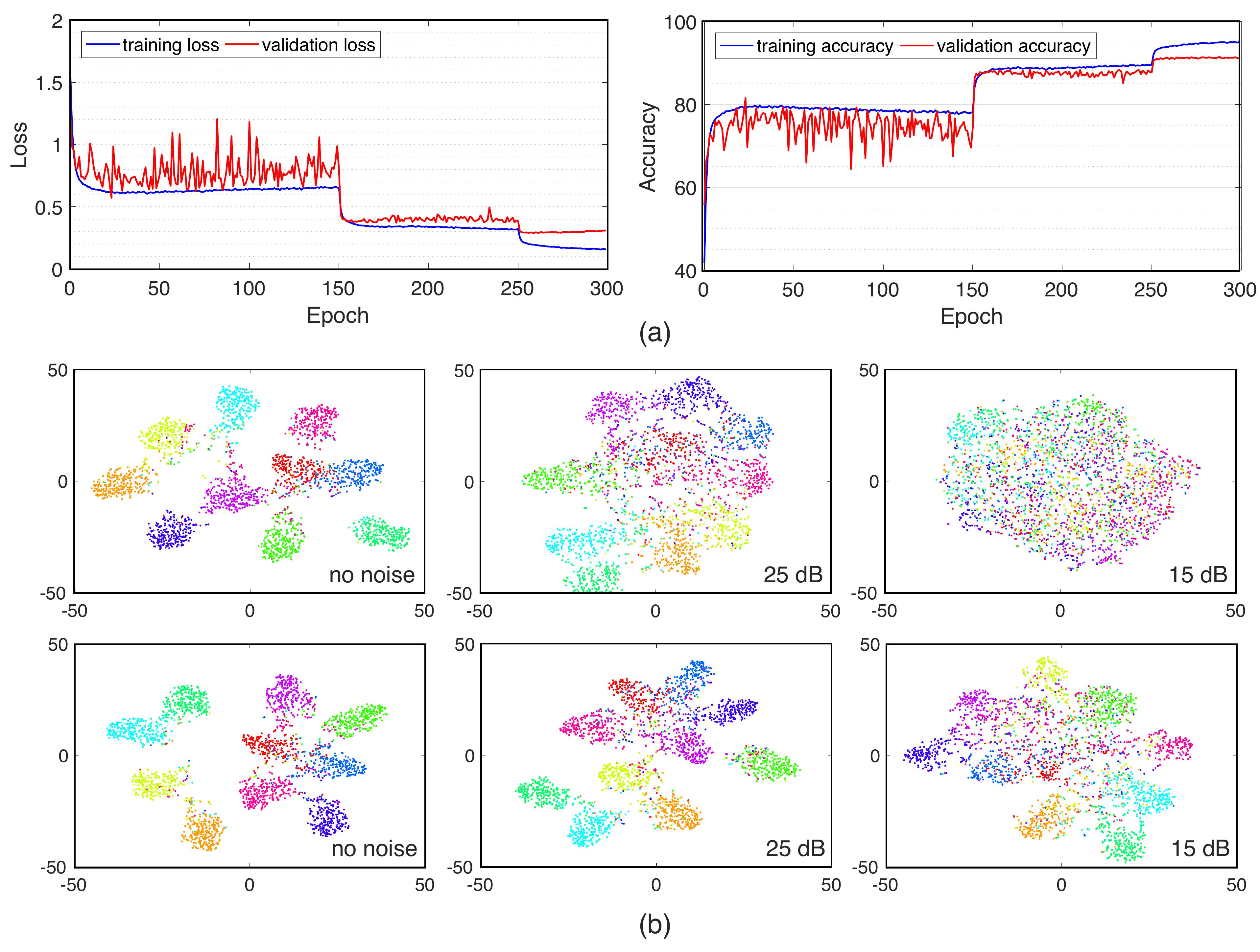}}
\caption{(a) Training and validation loss (left) and training and validation accuracy (right) of ScieNet with MobileNetV2; (b) embedding space visualization for baseline MobileNetV2 (top) and for ScieNet with MobileNetV2 (bottom) at different noise levels.}
\label{fig_vis}
\end{figure}

Fig.~\ref{fig_vis} (a) shows accuracy and loss during 300 epochs training process. The validation set contains 1,000 images separated from the inference set. To provide more insight of the performance difference between baseline and the proposed design, Fig.~\ref{fig_vis} (b) shows six embedding space visualizations of MobileNetV2 and ScieNet with MobileNetV2. The embedding space is taken between the two fully connected layers. For the three different levels of input noise, it can be observed that baseline network's capability to separate classes distinctively at high dimensional space decreases quickly with increasing noise level. When input images are clean the vectors in embedding space distribute into ten clusters with little overlap. As SNR drops to 25 dB shapes of clusters are still maintained but the distance between them shrinks and overlap increases. At 15 dB the network fails to provide meaningful clustering. On the other hand, the proposed design is able to maintained good separation between mapped vectors for each class from no noise to 25 dB. At 15 dB its clustering capability is affected but the deterioration is much lower than baseline.

\paragraph{Accuracy with Rainy images}
As shown in Table.~\ref{table: rain}, for all baseline networks tested, when rain drops are present in images, classification accuracy is decreased. For MobileNetV2, which is the smallest network tested, accuracy drop is more significant than the other two baseline. Under heavy rain condition, MobileNetV2 loses more than 40\% accuracy, while ResNet and DenseNet lose around 20\%. This demonstrates that structured perturbation has negative impact on deep learning performance and should not be overlooked. For ScieNet, performance is improved over all three baselines for both light and heavy rain conditions. Similar to the noisy input condition, ScieNet implemented with MobileNetV2 shows most accuracy gain.

\begin{figure}[t]
\centerline{\includegraphics[width=0.95\textwidth]{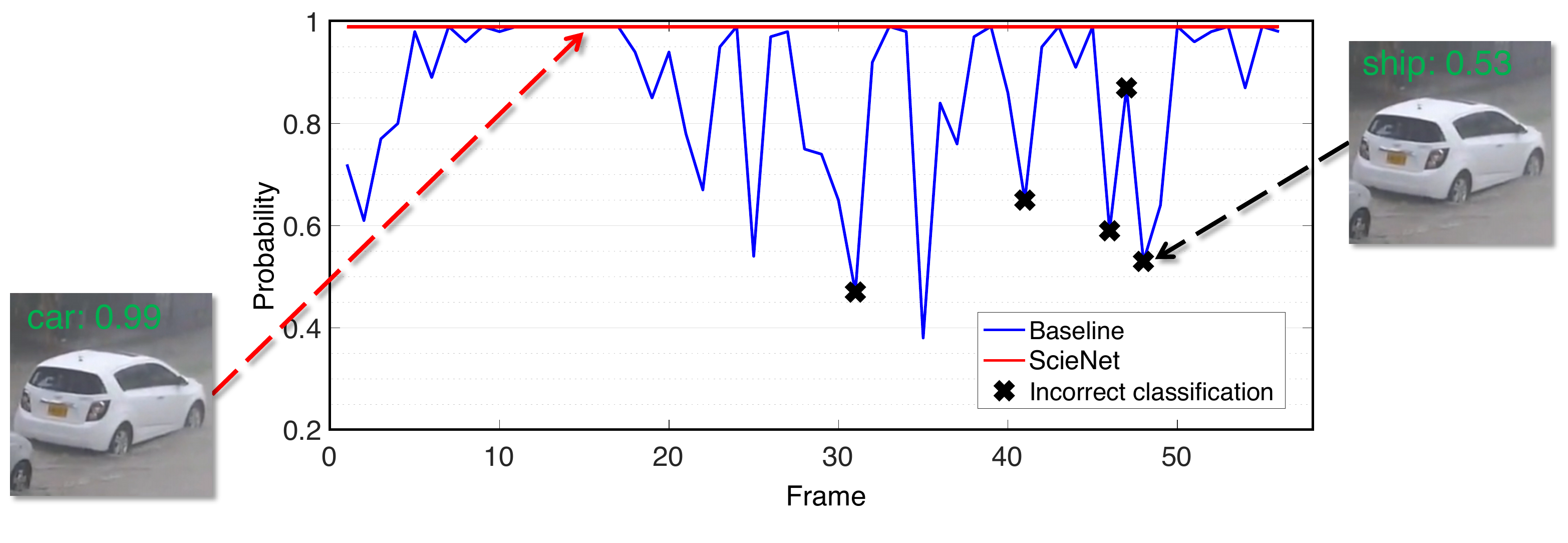}}
\caption{Probability of the top class in classification result for real-life rainy footage; cross sign marks frame that receive incorrect classification.}
\label{fig_rain}
\end{figure}

The real-life rainy footage we test contains a stationary car, and the comparison is between MobileNetV2 and ScieNet with MobileNetV2. Over all frames of the footage MobileNetV2 successfully classifies 87\% of them, while for ScieNet, 100\%. Probability of the top class from each network is shown in Fig.~\ref{fig_rain}. The probability from the baseline network varies over different frames of the footage and at several cases the network classify the object incorrectly as ship. ScieNet, on the other hand, differentiates its performance as its output is more stable: probability of the top class remains at 0.99 over all frames.

\begin{table}
  \caption{Accuracy (\%) results for rainy CIFAR10 images}
  \label{table: rain}
  \centering
  \begin{tabular}{llcc}
    \toprule
    Model    &  Clean &  Light rain & Heavy rain\\
    \midrule
    MobileNetV2          & 91.30  & 72.10    & 46.59  \\
    ResNet101            & 93.57  & 81.84    & 69.23    \\
    DenseNet121          & 93.00  & 81.72    & 67.70  \\
    \midrule
    \textbf{ScieNet with MobileNetV2} & 91.32 & 85.89 & 59.94  \\
    \textbf{ScieNet with ResNet101}    & 93.52 & 88.14 & 74.96  \\
    \textbf{ScieNet with DenseNet121}  & 93.07 & 87.50 & 75.01  \\
    \bottomrule
  \end{tabular}
\end{table}

\section{Conclusions}
We present a hybrid deep learning architecture called ScieNet, which is an integration of statistical machine learning and neuro-inspired learning models. It uses neuro-inspired learning to extract contextual information and DNN to perform classification on the context-space. The experimental results show that classification using ScieNet is resilient to stochastic (noisy) and structured (rainy) perturbation of inputs. More importantly, ScieNet demonstrates robust classification \textit{without the need for prior training on perturbed images}. In conclusion, ScieNet integrates neuro-inspired and statistical learning to create an attractive AI architecture for applications interacting with physical world, such as autonomous systems. Our future work will extend ScieNet to complex applications beyond image classification, for example, in decision-making, where imperfect input is the normal, not an aberration. 

\smallskip
\smallskip

\bigskip

\bibliographystyle{unsrt}
\bibliography{full_cite.bib}{}

\end{document}